\def\year{2021}\relax
\documentclass[letterpaper]{article} 
\usepackage{aaai21}  
\usepackage{times}  
\usepackage{helvet} 
\usepackage{courier}  
\usepackage[hyphens]{url}  
\usepackage{graphicx} 
\usepackage{natbib}  
\usepackage{caption} 
\usepackage{amsmath}
\usepackage{subfigure}
\usepackage{amsfonts}

\title{Reducing Neural Network Parameter Initialization Into an SMT Problem}
\author{

    Mohamad H. Danesh
    \\
}
\affiliations{

    Department of EECS, Oregon State University\\  Corvallis, OR, USA 97330\\
    
    daneshm@oregonstate.edu
}

\begin{document}

\maketitle

\begin{abstract}
Training a neural network (NN) depends on multiple factors, including but not limited to the initial weights. In this paper, we focus on initializing deep NN parameters such that it performs better, comparing to random or zero initialization. We do this by reducing the process of initialization into an SMT solver. Previous works consider certain activation functions on small NNs, however the studied NN is a deep network with different activation functions. Our experiments show that the proposed approach for parameter initialization achieves better performance comparing to randomly initialized networks.
\end{abstract}

\section{Introduction}
Satisfiability is the problem of determining if a formula has a model. In our case, formula is a set of propositions consisting of the weight and bias values of the NN and model is the set of initial values for them. Note that the described model is different from the common NN models. Intuitively, each possible model can be viewed as specifying a possible world within which a well formed formula can be evaluated \cite{barwise1977introduction}. Also, coming up with reasonable initial values for weights and biases of a NN is NP-complete \cite{Judd1990, Blum92traininga}. 

In this paper, we investigate the complexity of initializing parameters in a more complicated NN, with hidden layers, nonlinear activation functions, and on a complex task: classifying the hand written digits (MNIST). In this setting, initializing parameters is not an NP-complete problem anymore, but NP-hard. We tried to reduce our problem to have a framework that solves instances using an SMT-solver. 

\section{Approach} 
The NP-Hard problem we address answers the question: ``Is it possible to learn parameters of a deep NN for an arbitrary task such that it performs better than some relatively high threshold compared to randomly initialized parameters?''. Deep learning algorithms involve optimization in many contexts. The input of the optimization problem is a dataset and an objective, and the output is a set of values for the weights.

\begin{figure}[t]
\centering
\includegraphics[width=0.9\columnwidth]{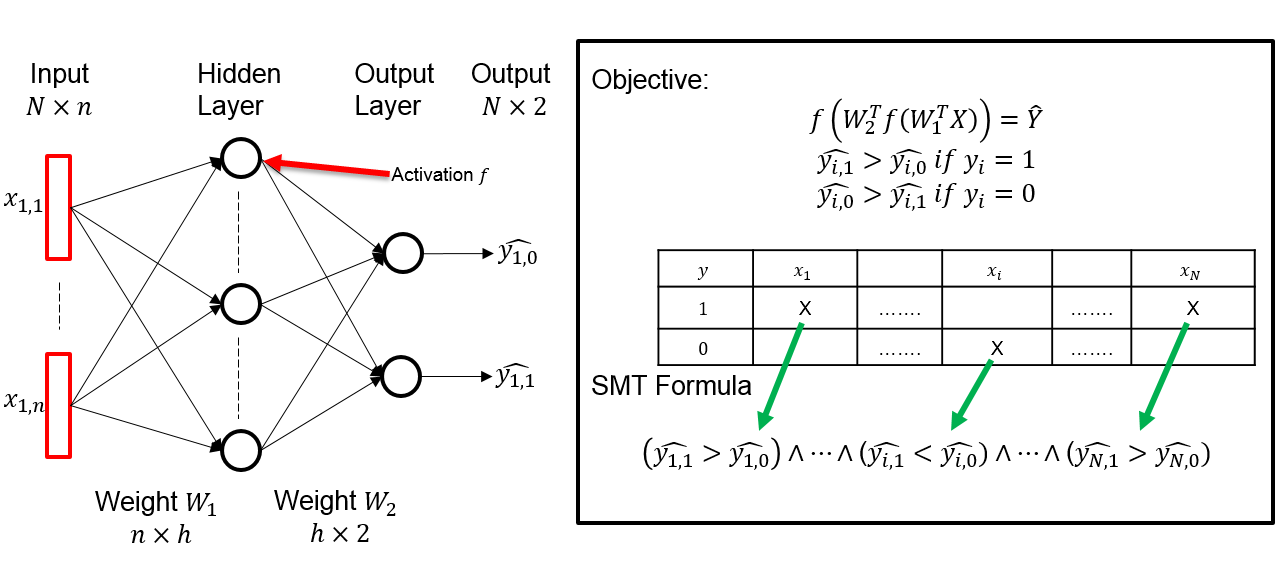}
\caption{Overall framework. Left: schematic demonstration of a DNN; Right: the equivalent SMT formulation.}
\label{fig:framework}
\end{figure}

Traditionally, parameters are initialized to small random values, and are updated across many iterations by using an optimization algorithm. This algorithm mostly uses gradient computation to find optimal values of parameters minimizing or maximizing an objective function.

If the weights of a NN are initialized to all $0$s, then the activation of each node will be $0$ as well. They will also all compute the same gradients during backpropagation and undergo the exact same parameter updates. In other words, there is no source of asymmetry between neurons if their weights are initialized to have the same values. 

Another approach is random initialization \cite{Glorot10understandingthe}. This way, weights are all random and unique in the beginning, so they compute distinct updates and integrate themselves as diverse parts of the full network. However, the problem is that there is no guarantee that model converges to an optimal weight assignment in a limited time frame. To converge faster, initialization of parameters is important. In this work, we investigate if SMT solvers could achieve a reasonable parameter initialization with a guarantee to better and faster convergence. First, the training process of a NN is reduced into an SMT problem for a binary classification problem. Second, the problem of integrating a non-linear activation function in an SMT solver and scalability of a large training set is investigated. Finally, the training results between randomly initialized weights and weights initialized by the results of an SMT solver are compared. 
\section{Reducing Training To SMT problem}
\subsection{NN Training}
We are given a dataset composed of $N$ input vectors $\{x_1,...,x_N\}$ of dimension $n$ and their respective labels $\{y_1,..., y_N\}$. $X$ has the dimension of $n \times N $ where each column corresponds to an input $x_i$, and similarly, $Y$ is the label matrix of dimension $N \times 2$. Each label is a one-hot encoded vector where $1$ at index zero means the sample is from class $0$ and at index one means the sample is from class $1$.

An NN computes an estimation of a label given an input sample. Inference is done according to the value of the activation output after the final layer of the network. We define $h$ as the network function and $\theta$ as the set of network weights. For the input $x_i$, we have the estimated output $h(x_i, \theta) = \hat{y}_i$ and the goal is to have $\hat{y}_i = y_i$. For a deep network composed of $k$ weights: $\theta = \{W_1,...,W_k\}$ and for an activation function $f$, the output can be expressed as:
\begin{equation}
     \hat{y}_i = h(x_i; \theta) = f(W_k^T f(W_{k-1}^T f(... f(W_1^T x_i)))
\end{equation}
Having the predicted output and the ground truth, the following formula, called binary cross-entropy, is typically used as a loss function for classification tasks:
\begin{equation}
     loss = -{(y\log(p) + (1 - y)\log(1 - p))}
\end{equation}
where $p$ is the predicted probability of observed output and $y$ is a binary value indicating if the class label is correct.

Each weight has an arbitrary number of connections which define together the overall architecture of a deep NN. We call $\{h_1, ..., h_{k-1}\}$ the value of these hidden nodes, in definitive: $W_1 \in \mathbb{R}^{n\times h_1}$, $W_2 \in \mathbb{R}^{h_1\times h_2}$, ..., $W_k \in \mathbb{R}^{h_{k-1}\times 1}$. Typically, a non-linear activation function $f$ is used to bring non-linearity in the NN function $h$. One of the most used functions is the ReLU function: 
\begin{equation}
f(x) = \max(0,x)
\label{eq:sig}
\end{equation}
The training is done in mini-batches of a certain size and is parameterized primarily in order to deal with large training datasets. The optimization algorithm is then used over many iterations organized in epochs to update the weights. It mostly consists of finding a local optima in the objective function and uses the backpropagation algorithm to compute the gradient of all parameters composing the network. 
 
 
\subsection{SMT Formula}  
Input of an SMT solver uses a set of variables representing real numbers that are expressed in classical order logic formula with equalities and/or inequalities, and translates it into a traditional $SAT$ formula. We want to express the task of NN with a formula where the input variables are the weights of the network. Since the weights represent the variables input to the SMT solver, we have in total $n \times h_1 \times ... \times h_{k-1}$ variables plus the corresponding bias terms.

To be consistent, we need the same settings given as input such as a training dataset, an architecture and a task. The architecture needs to be defined beforehand and will be integrated in the first-order logic formula. However, since we do not use the same algorithm for training as the classical approach, we do not use some hyper-parameter such as mini-batch size, learning-rate and number of training epochs.

We infer label $0$ if the first dimension of the output activation is higher than the second dimension as illustrated in Figure \ref{fig:framework}. We then express each clause of the SMT formulation as a part of the objective of the classification problem. The objective for one input $x_i$ is to have $\hat{y_i} = y_i$ which is expressed in as $\hat{y_{i,0}} > \hat{y_{i,1}}$ if $y_i=0$ and $\hat{y_{i,1}} > \hat{y_{i,0}}$ if $y_i=1$. A simplified version of the entire formula can be written as:
\begin{equation}
     (\hat{y_1} = y_1) \wedge ...  \wedge (\hat{y_N} = y_N)
\end{equation}
We notice that the length of the formula depends on the number of inputs provided in the training set. The more input samples are presented, the more constrained are the assignments of weights. This property links directly to the more traditional machine learning approach using gradient descent which highly rely on the number of training data.

We also noticed that putting all weights to zero can yield a satisfiable SMT formula depending if we use strict or loose inequalities. To avoid a dummy assignment of weights, another set of constraints added to enforce the values of the weights to be other than zero. We call $W_{ij}^{(k)}$, the value at the $i$th row and $j$th column of the weight at layer $k$. The formula then becomes:
\begin{equation}
\resizebox{.9\hsize}{!}{$(\hat{y_1} = y_1) \wedge ...  \wedge (\hat{y_N} = y_N) \wedge (W_{00}^{(0)} != 0) \wedge ... \wedge (W_{(k-1)2}^{(k)} != 0)$}
\end{equation}
The additional set of constraints is inspired from the dropout technique \cite{srivastava14a} used in NNs, where a layer sets a weight value to zero with a Bernoulli probability. In our case, we use the same method but instead we enforce random weight values to be different than zero.

One of the main components of NNs is the non-linear activation function. With that, it is possible to implement any function using NNs. To address this feature in the SMT formula, we set the activation function to ReLU (\ref{eq:sig}).

\bibliography{aaai21}

\end{document}


\section*{Supplementary Material}

\section{Related Work}
There have been multiple efforts to combine SAT/SMT solvers and neural networks in order to check the safety of neural networks and verifying them \cite{pulina2011checking, wang2018formal, katz2017reluplex, Pulina2012ChallengingSS}, or increase their accuracy \cite{arnault2019Precision}.

\cite{Judd1990} shows that by having a neural network and some data, finding a set of weights so that the neural network predicts the output correctly is NP-Complete. They also show that even predicting the output correctly for two-third of the training data is NP-Complete. So as a result, in the worst-case, it has been implied that training a neural network is difficult in its nature. \cite{Blum92traininga} investigate a simple neural network consisting of a 2-layer 3-node neural network with linear activation functions on a simple AND gate imitation task. They show that it is NP-Complete to find a set of weights so that the network produces output consistent with a given set of training examples. Their results imply that it is not possible to bypass computational difficulties by only using simple network architectures.

\cite{pulina2011checking} and \cite{Pulina2012ChallengingSS} suggest to combine SMT solvers and neural networks in order to verify neural networks. They have considered a multi-layer perceptron (MLP) for the verification. MLPs are considered a simple variation of neural networks, but are able to approximate most of the non-linear functions. They evaluate two types of safety conditions. One is to ensure that the output of the MLP is always in a threshold bound of the correct output, and the other one is that the output of the MLP is close to some known value or range of values modulo the expected error variance. \cite{katz2017reluplex} consider a deep neural network instead of an MLP. They verify deep NNs based on the simplex method which is extended to handle rectified linear unit (ReLU) activation function in recent neural networks. In this, work verification is done by looking at the neural network as a whole, rather than making simplifying assumptions. DNNs' verification is a cumbersome task since they have many parameters embedded in them, they are non-linear and non-convex. Thus, verifying DNNs is an NP-Complete task. In this work, in order to make verification feasible, they only use ReLU activation function in their DNN architecture. \cite{arnault2019Precision} look after the accuracy of the predictions made by a neural network. They have used both ReLU and \textit{tanh} activation functions in their neural network architecture. \cite{wang2018formal} apply internal arithmetic to bound the DNNs prediction outputs. They show that ReluVal gives a better performance comparing to Reluplex.

\section{Experiments}
\subsection{Analysis of SMT Solver Method}
In this section, we analyze the performance of the SMT solver ran over specific settings. We study how the number of input samples and the architecture used to build the SMT formula influence the performance and the running time. 

\begin{figure}[h]
\begin{center}
\subfigure[]{
\includegraphics[width=0.3\textwidth]{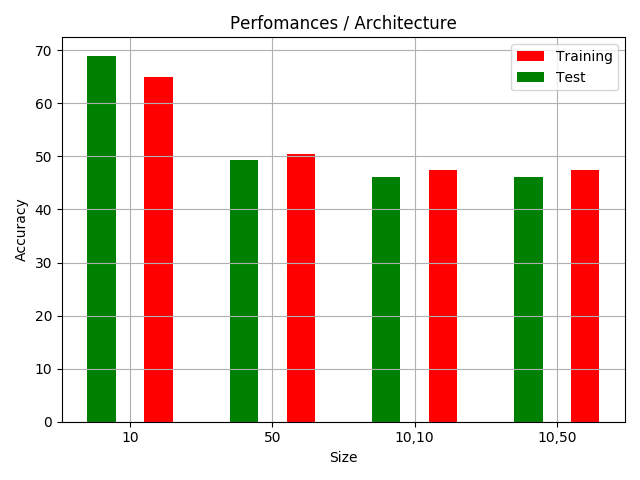}
}\hspace{0.01\columnwidth}
\subfigure[] {
\includegraphics[width=0.3\textwidth]{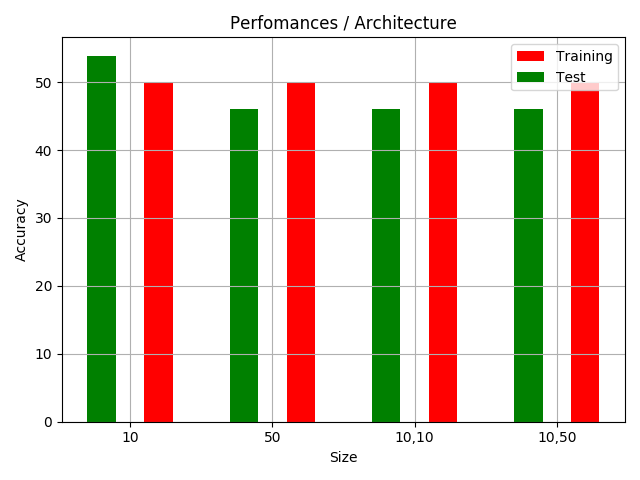}
}\hspace{0.01\columnwidth}
\subfigure[] {
\includegraphics[width=0.3\textwidth]{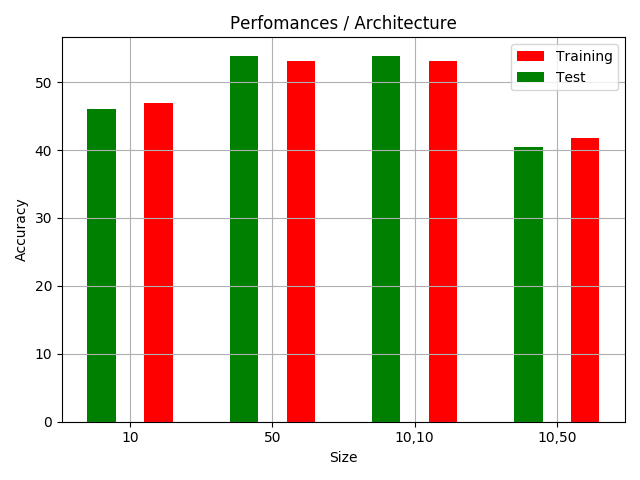}
}\hspace{0.01\columnwidth}
\subfigure[] {
\includegraphics[width=0.3\textwidth]{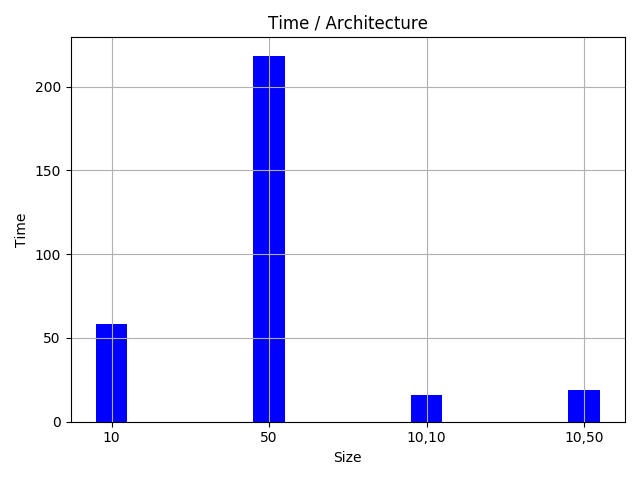}
}\hspace{0.01\columnwidth}
\subfigure[] {
\includegraphics[width=0.3\textwidth]{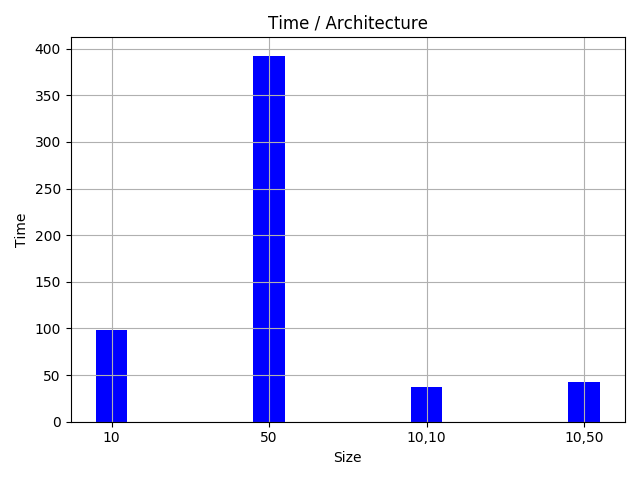}
}\hspace{0.01\columnwidth}
\subfigure[] {
\includegraphics[width=0.3\textwidth]{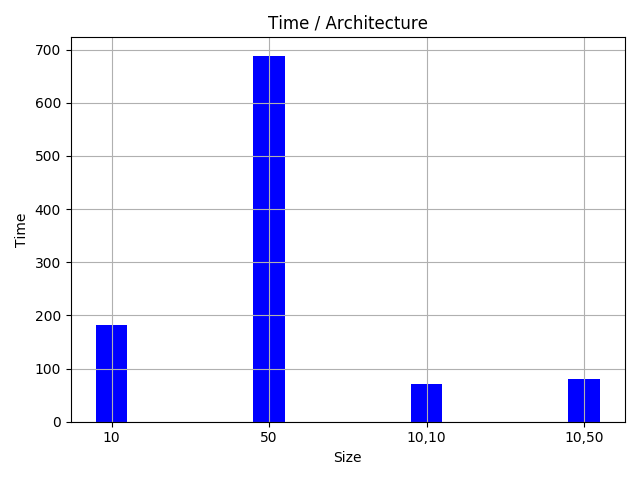}
}\hspace{0.01\columnwidth}
\end{center}
\caption{\small Experiments are done on different subset sizes of the dataset and show experiments on different network architectures. Different histograms are displayed and show the training accuracy (red) and validation accuracy (green) together (a), (b), (c) and the running time (blue) (d), (e), (f). The number of samples are $200$ for (a) and (d), $500$ for (b) and (e) and, $1000$ for (c) and (f). The horizontal axis describes what kind of architecture is used for the network.}
\label{fig:histograms}
\end{figure}

We also investigate the influence of the architecture on the running time and the performance given a certain amount of training data. Figure \ref{fig:histograms} shows different results for $3$ different sets of training data with different sizes, organized in columns: $200$, $500$ and $1000$, from left to right. We used $4$ different types of architecture for our experiments: $2$ architectures with one hidden layer (with $10$ and $50$ hidden nodes) and $2$ architectures with two hidden layer (with $10$ and $10$ then $10$ and $50$ hidden nodes per hidden layer). The first row (Figure \ref{fig:histograms}(a),(b),(c)) shows the accuracy for the subsampled training data and the entire validation set. The second row (Figure \ref{fig:histograms}(d),(e),(f)) shows the running time taken by the solver to assign values for the weights for each architecture.

These results show that the number of nodes in the first hidden layer has a high impact on the running time. By increasing the size of the first hidden layer, more computation is necessary and thus more time is used by the solver to output an assignment of values for the weights. We notice that the performances are not getting better when the architecture is more complex for a small training set. For a larger training set (Figure \ref{fig:histograms}(c),(f)), the best performances are achieved when the architecture is composed of one hidden layer with $50$ hidden nodes. We also experimented this training process over the entire training set ($4888$ input data) for an architecture composed of one hidden layer with $10$ hidden nodes and got $46.5$\% of accuracy for both the training and validation set, the solver used $1357$ seconds to output the result.

\begin{figure}[!th]
\begin{center}
\subfigure[] {
\includegraphics[width=0.2\textwidth]{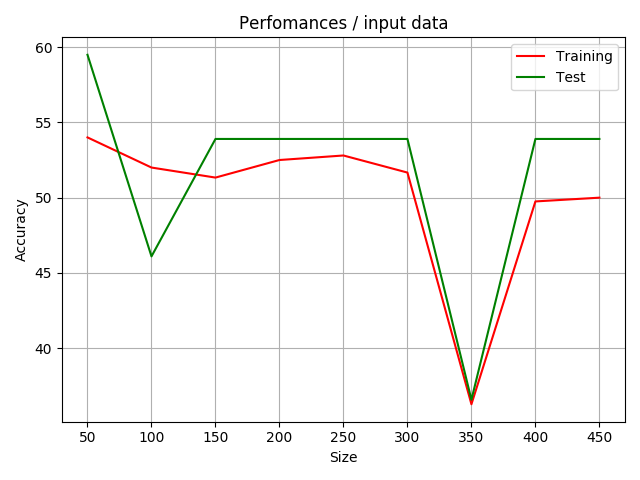}
}
\subfigure[] {
\includegraphics[width=0.2\textwidth]{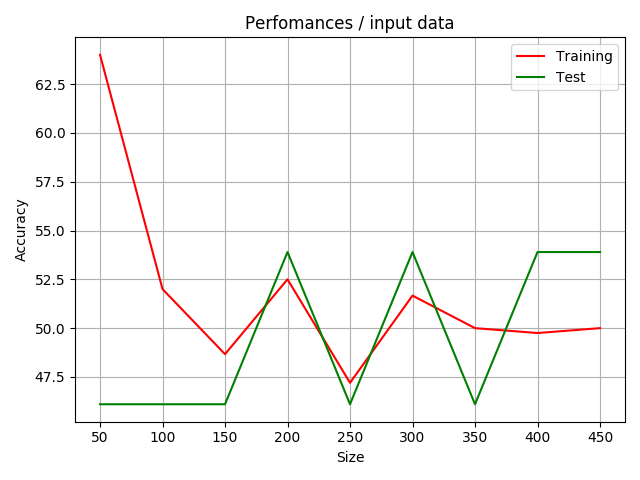}
}
\subfigure[] {
\includegraphics[width=0.2\textwidth]{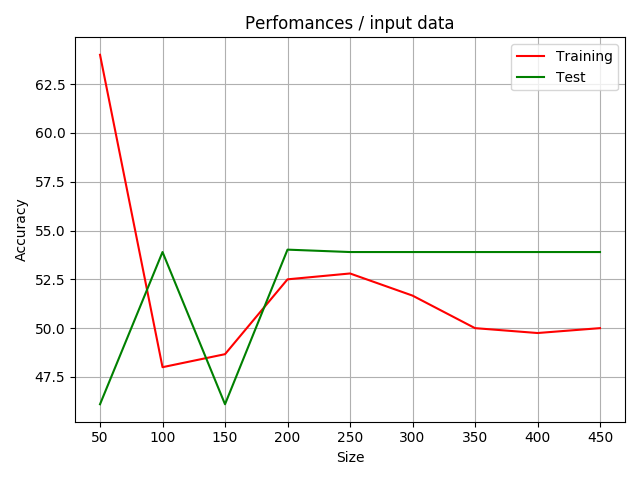}
}
\subfigure[] {
\includegraphics[width=0.2\textwidth]{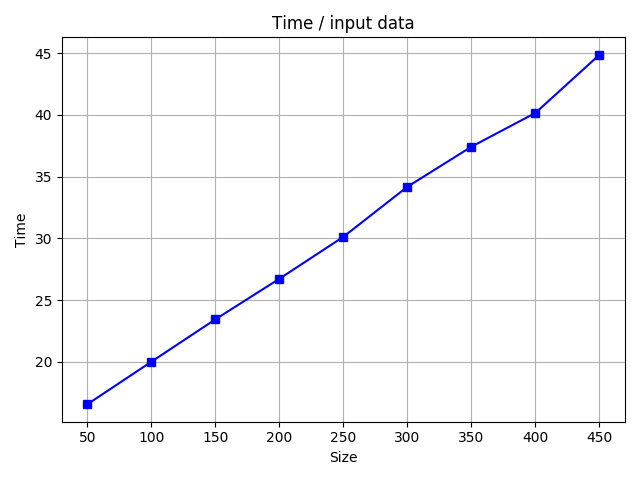}
}
\subfigure[] {
\includegraphics[width=0.2\textwidth]{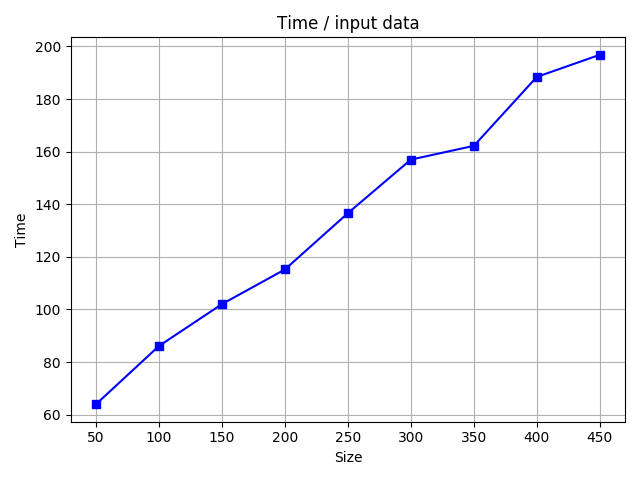}
}
\subfigure[] {
\includegraphics[width=0.2\textwidth]{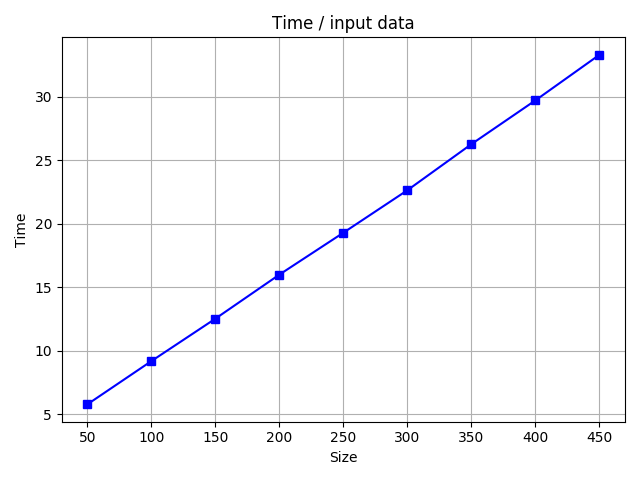}
}
\subfigure[] {
\includegraphics[width=0.2\textwidth]{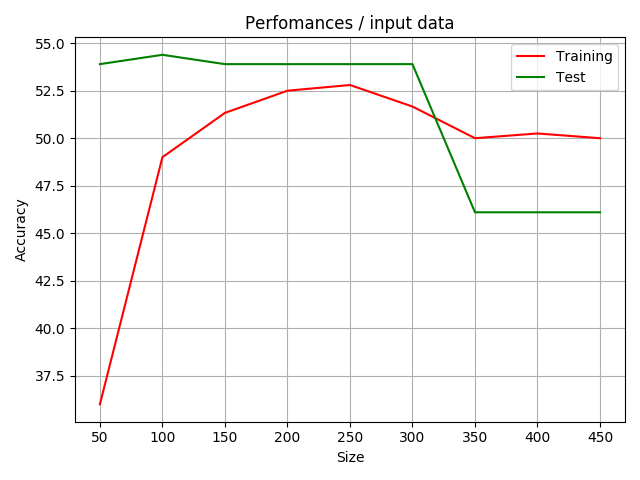}
}
\subfigure[] {
\includegraphics[width=0.2\textwidth]{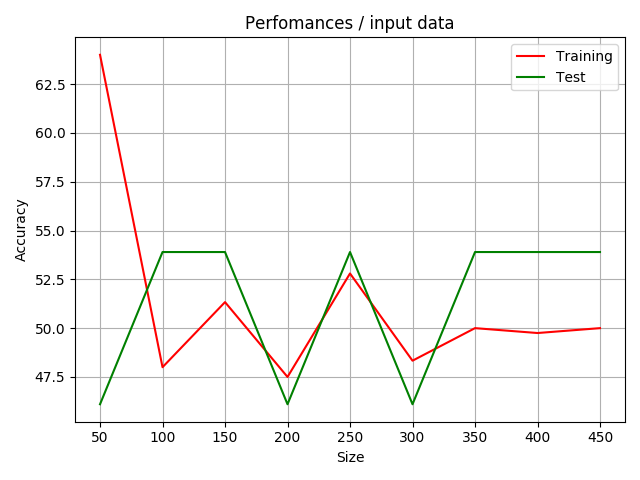}
}
\subfigure[] {
\includegraphics[width=0.2\textwidth]{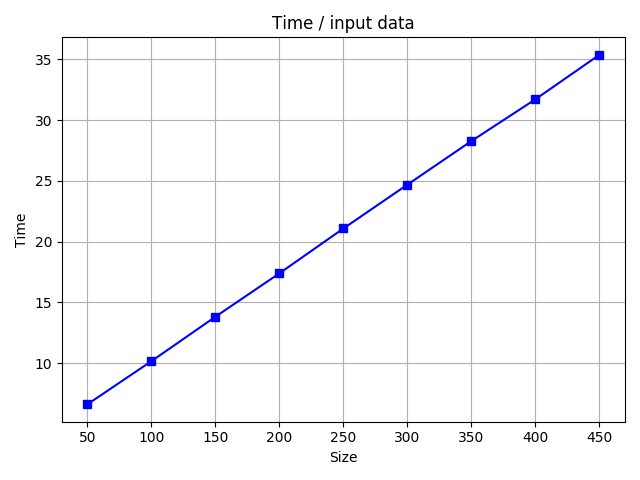}
}
\subfigure[] {
\includegraphics[width=0.2\textwidth]{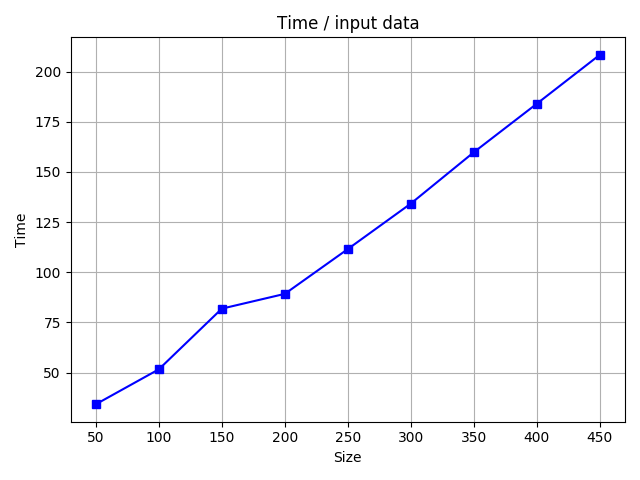}
}
\end{center}
\caption{\small Experiments done on different subset sizes of the dataset for different architectures. Showing training accuracy, validation accuracy and running time. (a), (b), (c), (g) and (h) shows the accuracy measurement for the architecture of size $(10)$, $(50)$, $(10,10)$, $(10,50)$ and $(50,50)$ respectively. Similarly, (d), (e), (f), (i) and (i) show the running time.
}
\label{fig:time}
\end{figure}

The second part is aimed at analyzing the influence of the number of training samples on the running time. Figure \ref{fig:time} shows different accuracy measurement and running time curves organized by column of size $2$ where each tuple represents a certain architecture. The architecture represented from top left to bottom right are: $(10)$, $(50)$, $(10,10)$, $(10,50)$ and $(50,50)$. The values represent the number of hidden nodes per hidden layer and multiple values mean multiple hidden layers. We primarily use a small amount of data and iteratively increase it, while recording the accuracy and time at each step. Figures \ref{fig:time}(d),(e),(f),(i),(e) display the running time and Figures \ref{fig:time}(a),(b),(c),(g),(h) display the performances both depending on the training set size.

From these figures, we can clearly see that the running time increases linearly in terms of the training subset size in all cases. As previously observed in the figure \ref{fig:histograms}, we also notice that the formula with a network having an architecture with a high number of hidden nodes on the first layer takes longer to solve (\ref{fig:time}(b),(h)). These results do not provide high performance in terms of accuracy, which means that the SMT solver did not find an optimal assignment of weights satisfying the input formula. However, the running time behaves as expected: it is dependant of the number of input samples fed into the formula.

\subsection{Weight Initialization Results}
This section compares the performance of a neural network which weights were computed using an SMT solver with a classic neural network which used random weight initialization.

\begin{figure}[!h]
\begin{center}
\subfigure[]{
\includegraphics[width=0.2\textwidth]{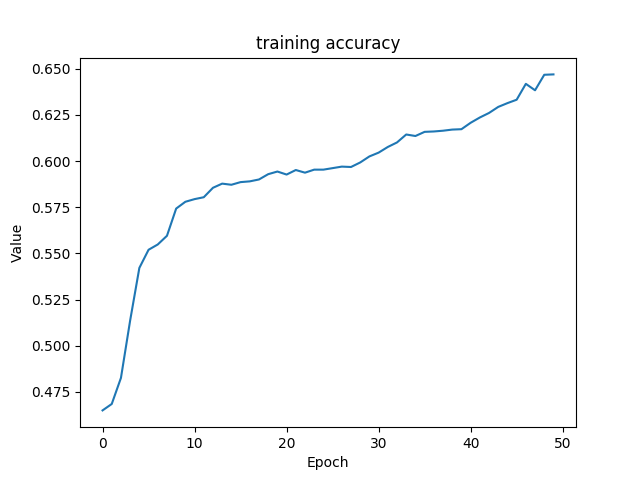}
}
\subfigure[] {
\includegraphics[width=0.2\textwidth]{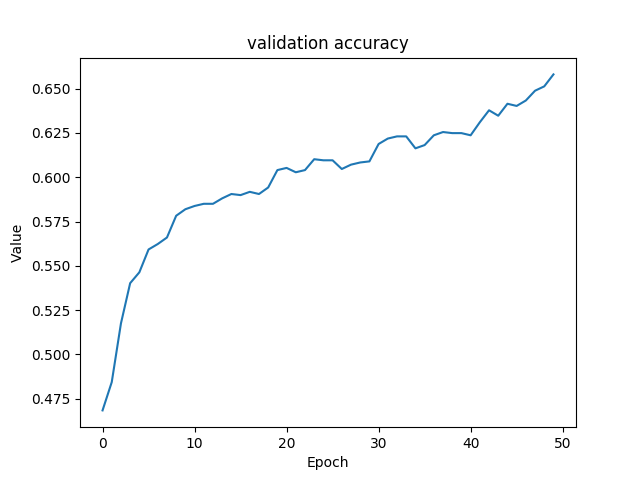}
}
\subfigure[] {
\includegraphics[width=0.2\textwidth]{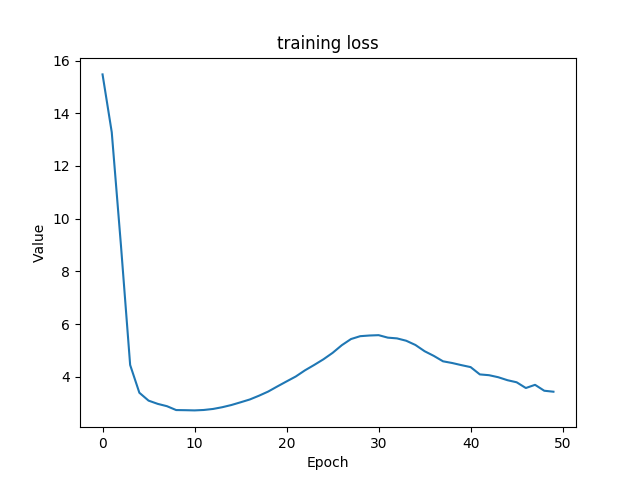}
}
\subfigure[] {
\includegraphics[width=0.2\textwidth]{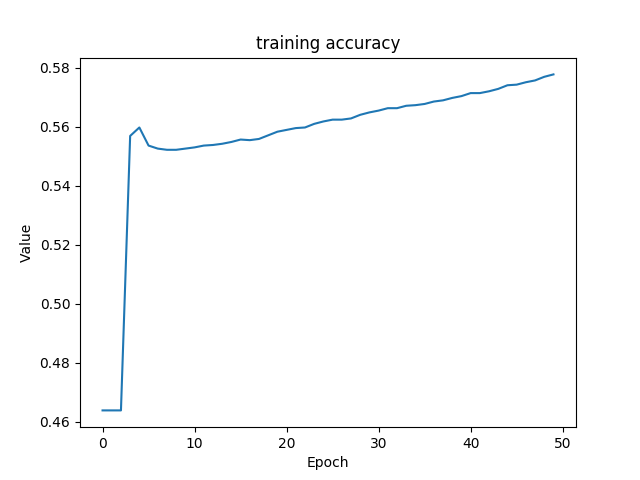}
}
\subfigure[] {
\includegraphics[width=0.2\textwidth]{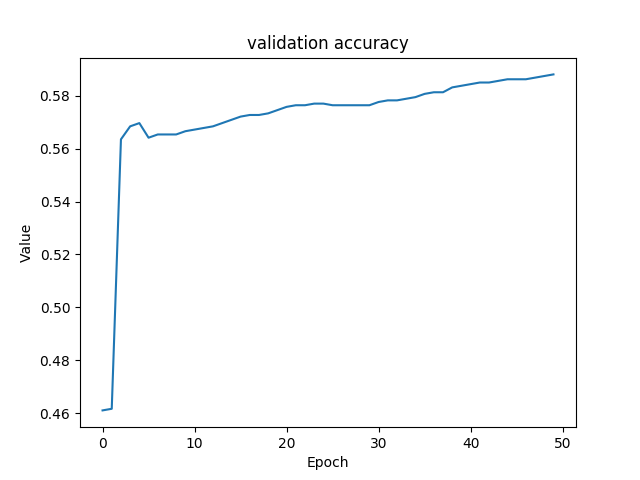}
}
\subfigure[] {
\includegraphics[width=0.2\textwidth]{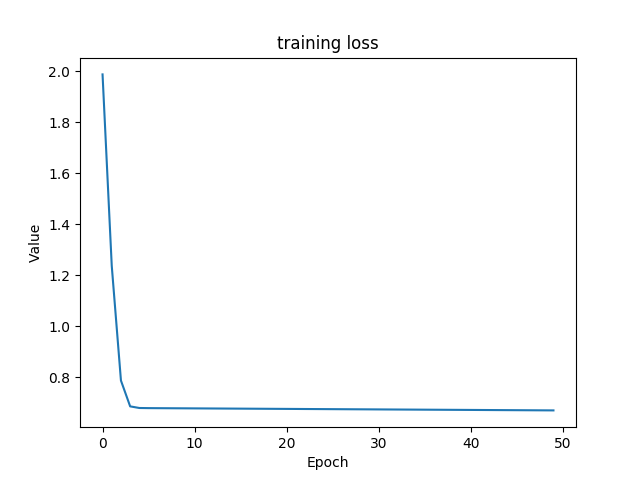}
}
\subfigure[] {
\includegraphics[width=0.2\textwidth]{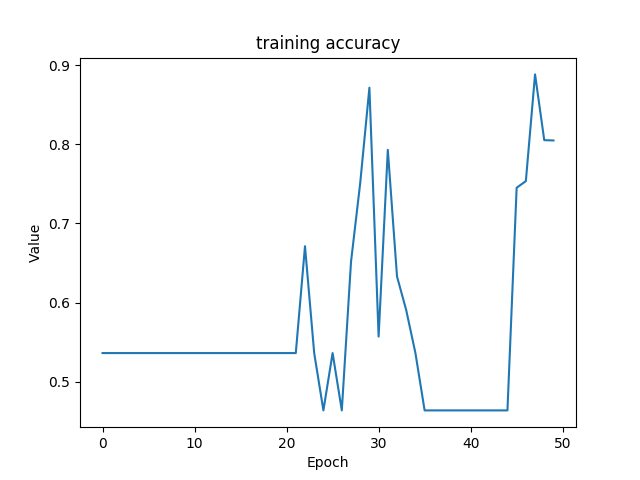}
}
\subfigure[] {
\includegraphics[width=0.2\textwidth]{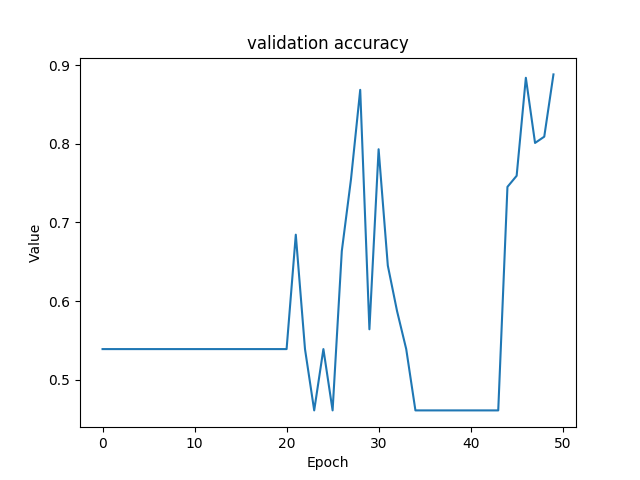}
}
\subfigure[] {
\includegraphics[width=0.2\textwidth]{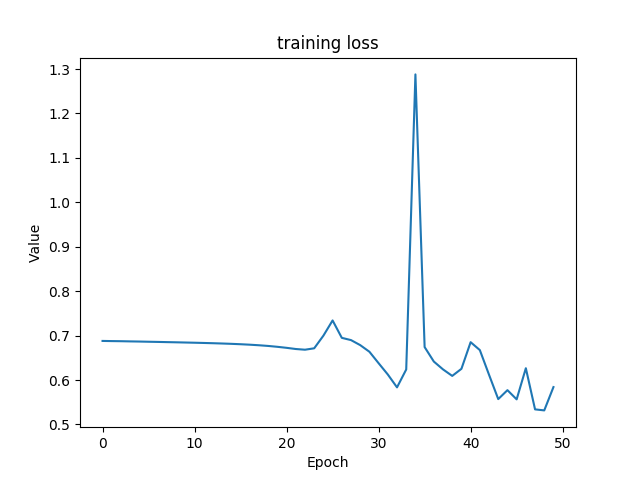}
}
\subfigure[] {
\includegraphics[width=0.2\textwidth]{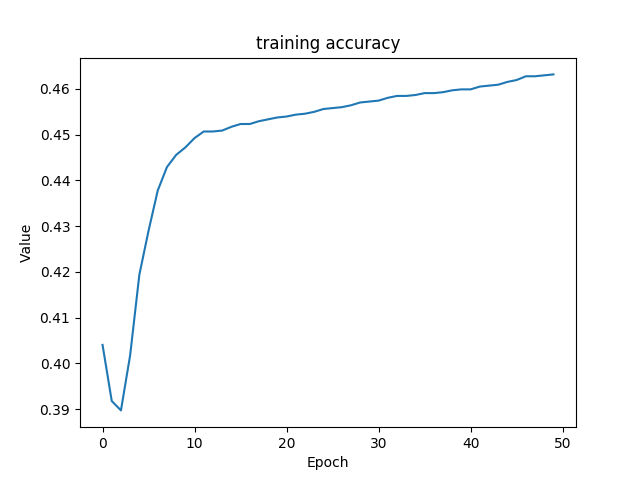}
}
\subfigure[] {
\includegraphics[width=0.2\textwidth]{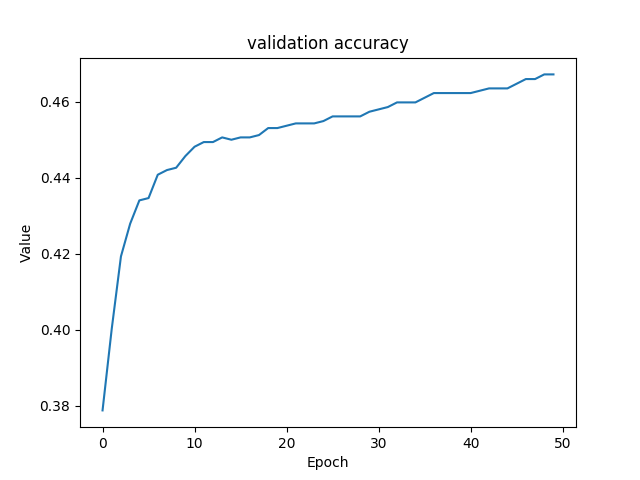}
}
\subfigure[] {
\includegraphics[width=0.2\textwidth]{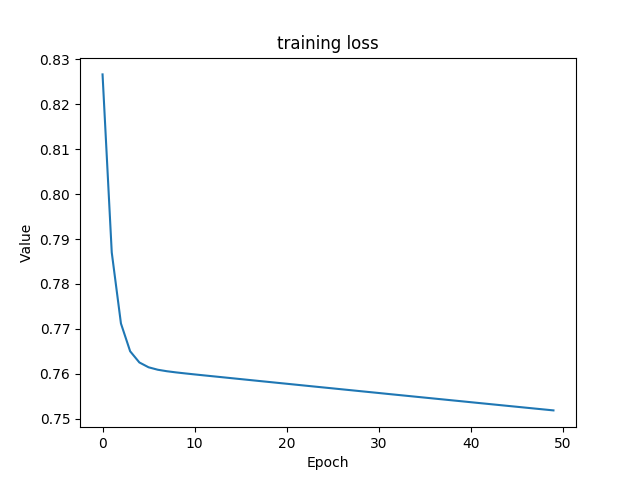}
}
\subfigure[] {
\includegraphics[width=0.2\textwidth]{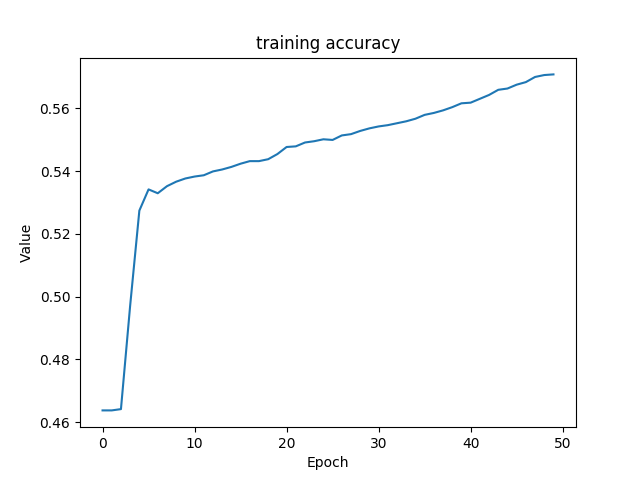}
}
\subfigure[] {
\includegraphics[width=0.2\textwidth]{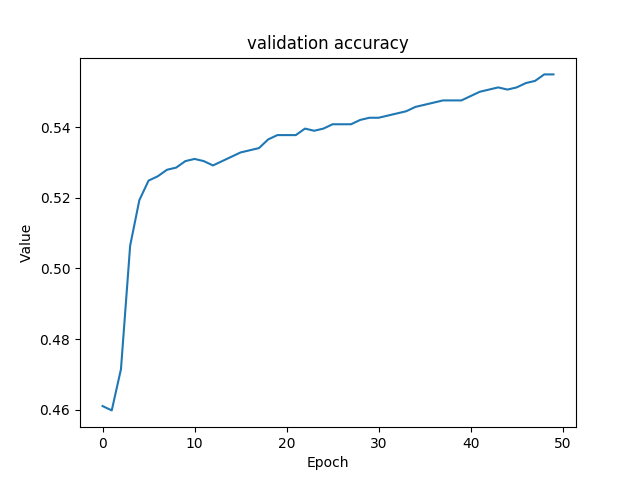}
}
\subfigure[] {
\includegraphics[width=0.2\textwidth]{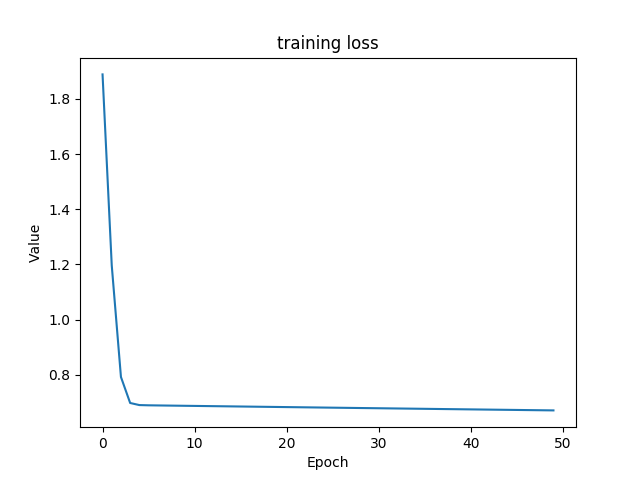}
}
\subfigure[] {
\includegraphics[width=0.2\textwidth]{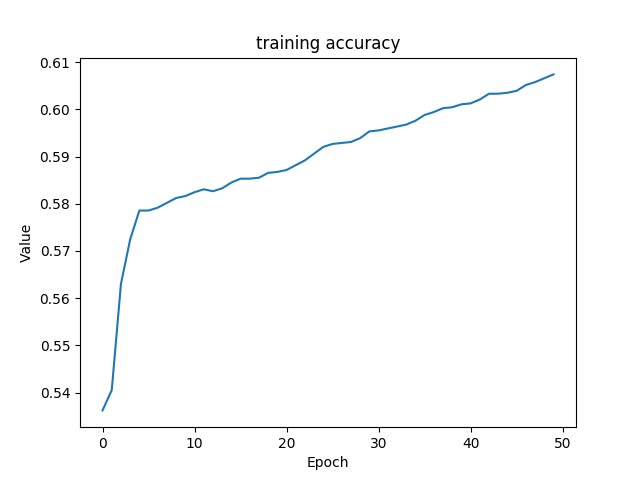}
}
\subfigure[] {
\includegraphics[width=0.2\textwidth]{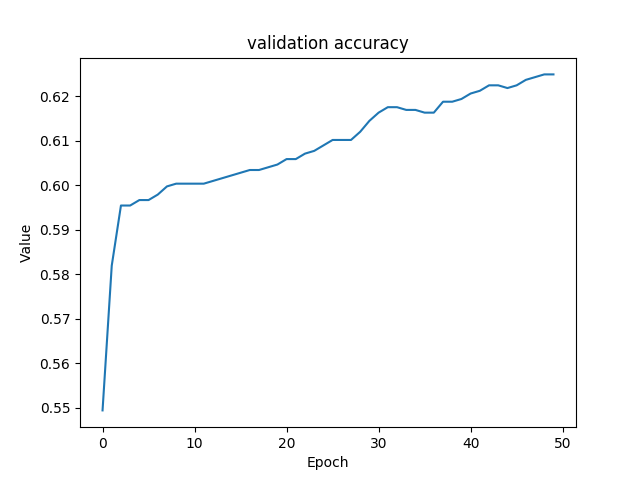}
}
\subfigure[] {
\includegraphics[width=0.2\textwidth]{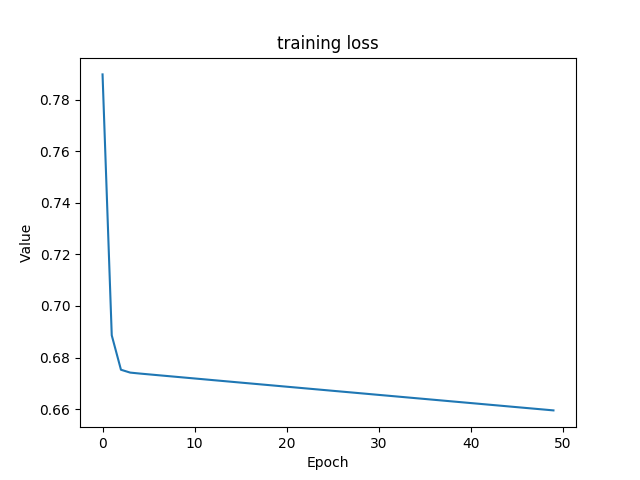}
}
\end{center}
\caption{\small Experiments done on the dataset. Showing training accuracy, validation accuracy and the training loss. (a), (b), (c) random initialization, and hidden layer of size 10. (d), (e), (f) SMT initialization, hidden layer of size 10, and 100 input samples to the SMT. (g), (h), (i) SMT initialization, two hidden layers of size 10, and 100 input samples to the SMT. (j), (k), (l) SMT initialization, hidden layer of size 50, and 100 input samples to the SMT. (m), (n), (o) SMT initialization, hidden layer of size 10, and 200 input samples to the SMT. (p), (q), (r) SMT initialization, hidden layer of size 10, and 500 input samples to the SMT. 
}
\label{fig:syn}
\end{figure}

The network is small regarding our dataset and do not have enough discriminative power to generalize. We chose to use such architecture to relax the computation complexity that will be necessary for the SMT solver to solve the problem. Figure~\ref{fig:syn} shows the actual training steps that were followed in order to get these prediction performances.

\begin{table*}[h]
\centering
\caption{Performance of the classification task}
 \begin{tabular}{| c | c | c| c| c| c|} 
 \hline
  Hidden layer & Initialization & SMT solver \# inputs & Loss & Validation Accuracy (\%)\\ [0.5ex] 
 \hline\hline
h=10 & random & - & 3.1 & 64.3 \\ 
h=10 & SMT & 100 & 0.75 & 59.5 \\
h=10,10 & SMT & 100 & \textbf{0.55} & \textbf{88.5} \\
h=50 & SMT & 100 & 1.92 & 46.2 \\
h=10 & SMT & 200 & 0.71 &  55.5 \\
h=10 & SMT & 500 & 0.67 & 62.4 \\

 \hline
\end{tabular}
\label{table:SGD_table}
\end{table*}

Table~\ref{table:SGD_table} shows the results for different settings and architectures of the neural network. The hidden layer column corresponds to the number of neurons in the hidden layer. As expected, by increasing the number of hidden layers, the performance of the network increases. In one experiment, by adding another hidden layer with the same size as the previous one to the network, which results in the best performance among other architectures. The initialization column shows how the weights were initialized. In one case, we use the random initialization, and in other cases we use our SMT solver results to initialize the weights. The next column shows the number of training set samples used as inputs to the SMT solver. It is expected that by increasing the number of samples given to the SMT solver, the performance of the network improves. The two last columns correspond to the loss and the validation accuracy of our network after training it for 50 epochs.

\label{fifff}

\section{Conclusion}
Briefly going through our work, we defined an approach to decide whether there exist weights and thresholds for the network so that it produces output consistent with a given set of training examples. Reducing neural network variable initialization to an SMT solver required careful and cumbersome work of defining a solid SMT formula. Solving the SMT formula gave us reasonable values for the weights and biases. Having the values, we did two different sets of experiments. First, we used them to construct a neural network, without any training using backpropagation. We compared the results with the classic neural network training with stochastic gradient descent. In this experiment, classic neural network performs better. Second, we used the values from SMT as the initial weights of the neural network and then applied stochastic gradient descent to that network. Comparing this setting to randomly initializing weights showed that using SMT to initialize weights performs better. However, the problem with SMT is to define a proper SMT formula so that we could get good initial values and it ends up costly in terms of running time.

\bibliography{aaai21}